\documentclass[runningheads]{llncs}

\usepackage[T1]{fontenc}
\usepackage{graphicx}

\usepackage{times}%
\usepackage{graphicx}
\usepackage{balance}  
\usepackage{hyperref}
\usepackage[utf8]{inputenc}
\usepackage{moreverb}
\usepackage{listings}
\usepackage{float}
\usepackage{graphicx}
\usepackage{adjustbox}
\usepackage{caption}
\usepackage{subcaption}
\usepackage{multirow}
\usepackage{xcolor}
\usepackage[ruled,linesnumbered,boxed]{algorithm2e}

\usepackage{algorithmic}
\usepackage{paralist}
\usepackage{amssymb}
\usepackage{amsmath}
\usepackage{todonotes}
\usepackage{pifont}
\usepackage{microtype}
\usepackage{xcolor}
\usepackage{colortbl}
\usepackage{amsmath}
\usepackage{dcolumn}
\usepackage{booktabs}
\usepackage{tikz}
\usepackage[misc,geometry]{ifsym}

\usepackage{paralist}
\usepackage{pgf-pie}
\usepackage{pgfplots}
\usepackage{txfonts}
\usepackage{orcidlink}
\usepackage{xcolor}
\newcommand{\uq}[1] {#1}

\newcommand{\real}{\mathrm{Re}}

\newcommand{\imag}{\mathrm{Im}}
\newcommand{\ConvolutionWithProjection}{\text{conv}}

\newcommand{\emb}[1]{\ensuremath{{\varphi}(#1)}}

\usepackage{amsmath,amssymb,amsfonts}

\begin{document}
\title{HybridFC: A Hybrid Fact-Checking Approach for Knowledge Graphs}
%
%

\author{Umair Qudus$^{(\textrm{\Letter})}$\orcidlink{0000-0001-6714-8729} \and
Michael R{\"o}der\orcidlink{0000-0002-8609-8277} \and
Muhammad Saleem\orcidlink{0000--0001-9648-5417} \and
Axel-Cyrille~Ngonga~Ngomo\orcidlink{0000-0001-7112-3516} 
}
\institute{DICE Group, Department of Computer Science,\\ Universit\"at Paderborn, Germany\\
\email{\{umair.qudus,michael.roeder,axel.ngonga\}@uni-paderborn.de,\\\{saleem\}@mail.uni-paderborn.de}\\
\url{https://dice-research.org/}
}

\authorrunning{Qudus et al.}
\maketitle              
\begin{abstract}
We consider fact-checking approaches that aim to predict the veracity of assertions in knowledge graphs. 
Five main categories of fact-checking approaches for knowledge graphs have been proposed in the recent literature, of which each is subject to partially overlapping limitations. In particular, current text-based approaches are limited by manual feature engineering. Path-based and rule-based approaches are limited by their exclusive use of  knowledge graphs as background knowledge, and embedding-based approaches suffer from low accuracy scores on current fact-checking tasks. We propose a hybrid approach---dubbed HybridFC---that exploits the diversity of existing categories of fact-checking approaches within an ensemble learning setting to achieve a significantly better prediction performance. In particular, our approach outperforms the state of the art by 0.14 to 0.27 in terms of Area Under the Receiver Operating Characteristic curve on the FactBench dataset. 
Our code is open-source and can be found at \url{https://github.com/dice-group/HybridFC}.

\keywords{fact checking  \and ensemble learning \and  knowledge graph veracity.}
\end{abstract}

\section{Introduction}
Knowledge graphs (KGs) are an integral part of the Web. A recent crawl of 3.2 billion HTML pages found over 82 billion RDF statements distributed over roughly half of the Web pages that were crawled.\footnote{\url{http://webdatacommons.org/structureddata/2021-12/stats/stats.html}} 
The increasing adoption of RDF at Web scale is further corroborated by the Linked Open Data cloud, which now contains over 10,000 KGs with more than 150 billion assertions and 3 billion entities.\footnote{\url{https://lod-cloud.net/}} 
Large-scale KGs like WikiData~\cite{stanislav2018wikidata}, DBpedia~\cite{auer2007DBpedia}, Knowledge Vault~\cite{knowledgevault2014Xin}, and YAGO~\cite{suchanek2007yago} contain billions of assertions, and describe millions of entities. They are being used as background knowledge in a growing number of applications, including healthcare~\cite{PUBHEALTH2020kotonya}, autonomous chatbots~\cite{dbpediaChatbot2018}, and in-flight entertainment~\cite{malyshev2018getting}. However, it is well established that current KGs are partially incorrect. For example, roughly 20\% of DBpedia's assertions are assumed to be false in the literature~\cite{defacto2015,tisco2019rula}. Fostering the further uptake of KGs at Web scale hence requires the development of highly accurate approaches that are able to predict the veracity of the assertions found in KGs in an automated fashion. We call such approaches fact-checking approaches.

%
In general, fact checking can be understood as the task of computing the likelihood that a given assertion is true~\cite{beyondFacts2021}. 
Various categories of automatic approaches have been proposed for this task. 
These categories include but are not limited to text-based~\cite{FactCheck2018,defacto2015}, path-based~\cite{coopal22019,KSREL2017,PRA2014,KL2015,pathcount2011},  rule-based~\cite{AMIE2013,AMIEplus2015,AMIE3}, and embedding-based~\cite{transE2013,transR2015} approaches. 
State-of-the-art instantiations of these categories of approaches are faced with a set of common limitations. In particular,
\begin{compactenum}[(1)]
\item Current text-based approaches rely on manual feature engineering~\cite{defacto2015,FactCheck2018,tisco2019rula}, which is time-consuming, and has been shown to be suboptimal w.r.t. their prediction performance by representation learning approaches~\cite{bengio2013representation}.
\item Path-based approaches rely on the availability of (short) paths in the KG between the entities that are part of the given assertion~\cite{coopal22019}.
\item Approaches that rely on KGs as background knowledge, i.e., path-, rule- and embedding-based approaches, have to take the open-world assumption (OWA) 
into account when determining the veracity of the given assertion~\cite{coopal22019}.
\item Embedding-based approaches~\cite{silva2021using} encounter limitations with respect to their accuracy~\cite{benchmark2018} as well as their scalability~\cite{kgesurvey2017}.
\end{compactenum}

We alleviate these limitations by exploiting the principles of diversity and accuracy known from ensemble learning. 
Our approach, dubbed HybridFC, overcomes the drawbacks of individual categories of approaches by leveraging the advantages of other categories of approaches. For example, we replace the manual feature engineering of the text-based approaches by exploiting embeddings.
To the best of our knowledge, we are the first to propose the combination of text-, path- and embedding-based fact-checking approaches in an ensemble learning setting.

The contributions of this work are as follows:
\begin{compactitem}
\item We use pre-trained KG embedding and sentence transformer models, 
and take advantage of transfer learning to reuse them for the task of fact checking.
\item We study the performance of different fact-checking approaches in isolation and in combination, and show that the joint use of multiple categories of approaches within an ensemble learning setting often leads to an improved performance.
\item We benchmark our approach on two recent fact-checking datasets, i.e., FactBench and BirthPlace/DeathPlace (BD). Our experiments suggest that our hybrid approach outperforms other text-, path-, rule- and embedding-based approaches by at least 
0.14 area under the curve (AUROC) on average on the FactBench dataset. It is ranked 3rd on the smaller BD dataset.
\end{compactitem} 

The rest of this paper is structured as follows. In Section \ref{sec:preliminaries}, we introduce the notation required to understand the rest of the paper. In Section \ref{sec:motivationRelatedWork}, we give related work and motivate our work using a real-world example. In Section \ref{sec:methodology}, we present HybridFC.  
Thereafter, the evaluation datasets and metric used are presented in Section \ref{sec:evaluation}. We then discuss our results in Section \ref{sec:results}. In Section~\ref{sec:ablation}, we present an ablation study of our approach. Finally, we conclude and discuss potential future work in Section~\ref{sec:conclusion}.

\section{Preliminaries}
\label{sec:preliminaries}
In this section, we define the terminology and notation used throughout this paper. 
We build upon the definition of fact checking for KGs suggested in~\cite{FactCheck2018}:
\begin{definition}[Fact Checking]
Given an assertion, a reference KG $\mathcal{G}$, and/or a reference corpus, fact checking is the task of computing the likelihood that the given assertion is true or false~\cite{FactCheck2018}.
\end{definition}
Throughout this work, we rely on RDF KGs:
\begin{definition}[RDF Knowledge Graph]
An RDF KG $\mathcal{G}$ is a set of RDF triples $\mathcal{G} \subseteq (\mathbb{E} \cup \mathbb{B}) \times \mathbb{P} \times (\mathbb{E} \cup \mathbb{B} \cup \mathbb{L})$, where each triple $(s,p,o) \in \mathcal{G}$ comprises a subject $s$, a predicate $p$, and an object $o$. $\mathbb{E}$ is the set of all RDF resource IRIs, $\mathbb{B}$ the set of all blank nodes, $\mathbb{P} \subseteq \mathbb{E}$ the set of all RDF predicates, and $\mathbb{L}$ the set of all literals~\cite{COPAAL2019}.
\end{definition} 
In our approach, we use multiple representations of RDF KGs. In addition to their representation as sets of assertions, we also exploit representations in continuous vector spaces, called embeddings~\cite{kgesurvey2017,kgesurvey2020}.  
\begin{definition}[KG Embeddings]
A KG embedding function $\varphi$ maps a KG $\mathcal{G}$ to a continuous vector space.
Given an assertion $(s, p, o)$, ${\varphi}(s), {\varphi}(p),$ and ${\varphi}(o)$ stand for the embedding of the subject, predicate, and object, respectively. 
Some embedding models map the predicate embedding into a vector space that differs from the space wherein ${\varphi}(s)$ and ${\varphi}(o)$ are mapped. 
For those models, we use ${\varphi}^*(p)$ to denote predicate embeddings.
\end{definition}

Different embedding-based approaches use different scoring functions to compute  embeddings~\cite{kgesurvey2017}. 
The approaches considered in this paper are shown in  Table~\ref{table:embdings}. 

\begin{table}[h!tb]
 \centering
 \caption{Scoring functions of different embedding-based approaches used in this paper. $\otimes$ stands for the quaternion multiplication, $\mathbb{R}$ for the space of real numbers, $\mathbb{H}$  for the space of quaternions, $\mathbb{C}$ for the complex numbers, $\real$ for the real part of a complex number, $\imag$ for the imaginary part of a complex number, $\ConvolutionWithProjection$ for the convolution operator, $\overline{\varphi(o)}$ for the complex conjugate of $\varphi(o)$, $q$ is the length of embedding vectors, $\cdot$ for the dot product and $\left\|\cdot\right\|_{2}$ for the L2 norm.}
 \label{tab:mapping}
 \begin{tabular}{@{}l@{\hspace{0.2cm}}l@{\hspace{0.2cm}}l@{\hspace{0.2cm}}l@{}}
 \toprule 
 Approach & Scoring function& VectorSpace& Regularizer \\
 \midrule
 TransE & $
\left\|\left(\varphi(s)+\varphi(p)\right)-\varphi(o)\right\|_{2}
$&$\varphi(s), \varphi(p), \varphi(o) \in \mathbb{R}^{q}$&L2\\
 ComplEx & $\operatorname{Re}\left(<\varphi(s), \varphi(p), \overline{\varphi(o)}>\right)$&$\varphi(s), \varphi(p), \varphi(o) \in \mathbb{C}^{q}$&Weighted L2\\
 QMult & ${\varphi}(s) \otimes {\varphi}(p) \cdot {\varphi}(o)$&${\varphi}(s), {\varphi}(p), {\varphi}(o) \in \mathbb{H}^{q}$ & Weighted L2\\
 ConEx & $\real(\langle \ConvolutionWithProjection(\emb{s},\emb{p}), \emb{s}, \emb{p}, \overline{ \emb{o}} \rangle)$&$\varphi(s), \varphi(p), \varphi(o) \in \mathbb{C}^{q}$&Dropout, BatchNorm\\
 \bottomrule
\end{tabular}
\label{table:embdings}
\end{table}

\begin{definition}[Sentence Embedding Model]
A sentence embedding model maps the natural language sentence $t$ to a continuous vector space~\cite{sbert2019Reimers}. 
Let $b$ be the embedding function and let $T=(t_1,\ldots,t_k)$ be a list of $k$ sentences. 
We create the embedding vector for $T$ by concatenating the embedding vectors of the single sentences.
\end{definition}



\section{Related Work}
\label{sec:motivationRelatedWork}
We divide the existing fact-checking approaches into 5 categories: text-based~\cite{FactCheck2018,defacto2015}, path-based~\cite{COPAAL2019,KSREL2017}, rule-based~\cite{AMIE2013,AMIEplus2015,AMIE3}, KG-embedding-based~\cite{transD2015,transE2013,transR2015}, and hybrid approaches~\cite{Facty2017,tracy2019,ExFact2019}. 
In the following, we give a brief overview of state-of-the-art approaches in each category along with their limitations.

\subsection{Text-based Approaches}
Approaches in this category validate a given assertion by searching for evidence in a reference text corpus. FactCheck~\cite{FactCheck2018} and DeFacto~\cite{defacto2015} are two instantiations of this category. 
Both approaches search for pieces of text that can be used as evidence to support the given assertion by relying on RDF verbalisation techniques. 
TISCO~\cite{tisco2019rula} relies on a temporal extension of DeFacto. All three approaches rely on a set of manually engineered features to compute a vectorial representation of the texts they retrieved as evidence. 
This manual feature engineering often leads to a suboptimal vectorial representation of textual evidence~\cite{bengio2013representation}. 
In contrast, we propose the use of embeddings to represent pieces of evidence gathered from text as vectors. 
First, this ensures that our approach is aware of the complete 
piece of textual evidence
instead of the fragment extracted by previous approaches. 
Second, it removes the need to engineer features manually and hence reduces the risk of representing text with a possibly suboptimal set of manually engineered features.

\subsection{Path-based Approaches}
Path-based approaches generally aim to validate the input assertion by first computing short paths from the assertion's subject to its object within the input KG. 
These paths are then used to score the input assertion. 
Most of the state-of-the-art path-based approaches, such as COPAAL~\cite{COPAAL2019}, Knowledge stream~\cite{KSREL2017}, PRA~\cite{PRA2014}, SFE~\cite{SFE2015efficient}, and KG-Miner~\cite{KGMiner2016} rely on RDF semantics~(e.g., class subsumption hierarchy, domain and range information) to filter useful paths. However, the T-Box of a large number of KGs provides a limited number of RDFS statements. Furthermore, it may also be the case that no short paths can be found within the reference KG, although the assertion is correct~\cite{COPAAL2019}. In these scenarios, path-based approaches fail to predict the veracity of the given assertion correctly.\footnote{For the assertion $award\_00135$ from the FactBench, COPAAL produces a score of $0.0$ as it is unable to find a path between the assertion's subject and its object.}

\subsection{Rule-based Approaches}
State-of-the-art rule-based models such as KV-Rule~\cite{unsupervised2020rule}, AMIE~\cite{AMIE2013,AMIEplus2015,AMIE3}, OP~\cite{ontoPathFining2016}, and RuDiK~\cite{rudik2018}  extract association rules to perform fact checking or fact prediction on KGs. To this end, they often rely on reasoning~\cite{AMIE3,rule2022}.  These approaches are limited by the knowledge contained within the KG, and mining rules from large-scale KGs can be a very slow process in terms of runtime~(e.g., OP takes $\ge 45$ hours on DBpedia~\cite{AMIE3}).

\subsection{Embedding-based Approaches}
Embedding-based approaches use a mapping function to represent the input KG in a continuous low-dimensional vector space
~\cite{transD2015,transE2013,transR2015,conexdemir2021convolutional,complextrouillon2016complex,trustworthyKG,silva2021using}. 
For example, Esther~\cite{silva2021using} uses compositional embeddings to compute likely paths between resources.  \uq{TKGC~\cite{trustworthyKG} checks the veracity of assertions extracted from the Web before adding them to a given KG. The veracity of assertions is calculated by creating a KG embedding model 
and learning a scoring function to compute the veracity of these assertions.}
In general, embedding-based approaches are mainly limited by the knowledge contained within the continuous representation of the KG. 
Therefore, these approaches encounter limitations with respect to their accuracy in fact-checking scenarios~\cite{benchmark2018} as well as their scalability when applied to large-scale KGs~\cite{kgesurvey2017}. 

\subsection{Hybrid Approaches}

While the aforementioned categories have their limitations, they also come with their own strengths.
\uq{Consider the assertion in Listing~\ref{lst:mv1}. The text-based approach FactCheck cannot find evidence for the assertion. A possible reason might be that West Hollywood is not mentioned on the Wikipedia page of Johnny Carson. 
However, COPAAL finds evidence in the form of corroborative paths that connect the subject and the object in DBpedia. For example, the first corroborative path in this particular example from FactBench~\cite{defacto2015} encodes that if two individuals share a death place, then they often share several death places. While this seems counter-intuitive, one can indeed have several death places by virtue of the part-of relation between geo-spatial entities, e.g., one's death places can be both the Sierra Towers and West Hollywood.  
In our second example shown in Listing~\ref{lst:mv2}, COPAAL is not able to find any relevant paths between the subject and the object. This shows one of the weaknesses of COPAAL which does not perform well for rare events, e.g., when faced with the \texttt{:award} property~\cite{COPAAL2019}.
In contrast, TransE~\cite{transE2013} is able to classify the assertion as correct.}
These examples support our hypothesis that there is a need for a hybrid solution in which the limitations of one approach can be compensated by the other approaches.

\begin{lstlisting}[frame=single, basicstyle=\fontsize{7}{7}\selectfont\ttfamily,
label={lst:mv1}, caption={\uq{Example 1 (correct, \texttt{death-00129.ttl} in the FactBench dataset~\cite{defacto2015}).}}]
PREFIX dbr:  <http://dbpedia.org/resource/>
PREFIX dbo:  <http://dbpedia.org/ontology/>
Assertion:  dbr:Johnny_Carson dbo:deathPlace dbr:West_Hollywood,_California

FactCheck Result: Score: 0.0
Proofs: [no proofs found]
========================================================
COPAAL Result: Score: 0.99
Proofs: evidence paths:[
evidence path 1: "predicate path: dbo:deathPlace/^dbo:deathPlace/dbo:deathPlace",
evidence path 2: "predicate path: dbo:deathPlace/^dbo:recordedIn/dbo:recordedIn",
...]
\end{lstlisting}

\begin{lstlisting}[frame=single, basicstyle=\fontsize{7}{7}\selectfont\ttfamily,
label={lst:mv2},caption={\uq{Example 2 (correct, \texttt{award-00135.ttl} in the FactBench dataset~\cite{defacto2015}).}}]
Assertion:  dbr:T._S._Eliot dbo:award dbr:Nobel_Prize_in_Literature

COPAAL Result: Score: 0.0
Proofs: evidence paths: [no paths found]
========================================================
TransE Result: Score: 0.90
\end{lstlisting}

FACTY~\cite{Facty2017}, ExFaKT~\cite{tracy2019}, and Tracy~\cite{ExFact2019} are hybrid approaches that exploit structured as well as textual reference knowledge to find the human-comprehensible explanations for a given assertion. ExFaKT and Tracy\footnote{https://www.mpi-inf.mpg.de/impact/exfakt\#Tracy} make use of rules mined from the KG. 
A given assertion is assumed to be correct if it fulfills all conditions of one of the mined rules. 
These conditions can be fulfilled by facts from the KG or by texts retrieved from the Web. 
The output of these approaches is not a veracity score. Rather, they produce human-comprehensible explanations to support human fact-checkers. \uq{Furthermore, these approaches are not designed for ensemble learning settings. They incorporate a text search merely to find support for the rules they generate. As such, they actually address  different problem statements than the one addressed herein.}
FACTY leverages textual reference and path-based techniques to find supporting evidence for each triple, and subsequently predicts the correctness of each triple based on the found evidence. 
Like Tracy and ExFaKT, FACTY only combines two different categories and mainly focuses on generating human-comprehensible explanations for candidate facts.  
To the best of our knowledge, our approach is the first approach that uses approaches from three different categories with the focus on automating the fact-checking task.

\begin{figure}[!htb]
\centering
\includegraphics[width=1\linewidth]{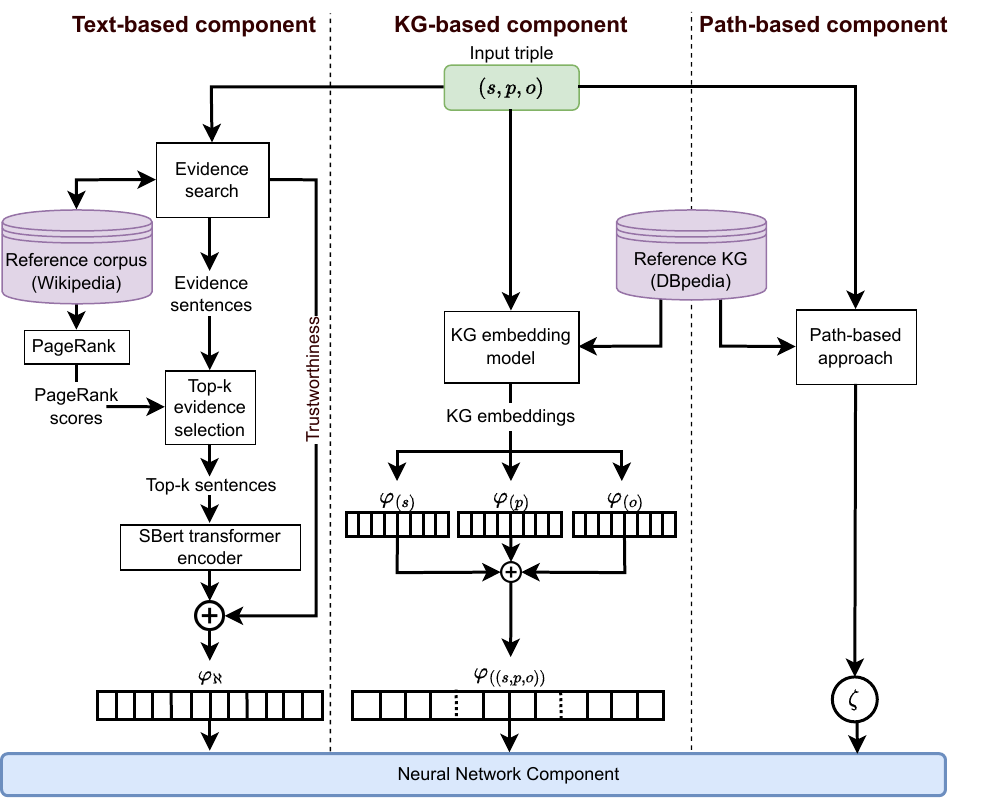}
\caption{Architecture of HybridFC. The purple color represents reference knowledge. The green color marks the input assertion. KG stands for knowledge graph. }
\label{fig:arch}
\end{figure}

\section{Methodology}
\label{sec:methodology}

The main idea behind our approach, HybridFC, is to combine fact-checking approaches from different categories. 
To this end, we created components for a text-based, a path-based and a KG embedding-based fact-checking algorithm. Figure \ref{fig:arch} depicts a high-level architecture of our  approach. 
We fuse the results from the three components and feed them into a neural network component, which computes a final veracity score. In the following, we first describe the three individual components of our approach in detail. Thereafter, we describe the neural network component that merges their results.

\subsection{Text-based Component}


Text-based approaches typically provide a list of scored text snippets that provide evidence for the given assertion, together with a link to the source of these snippets and a trustworthiness score~\cite{defacto2015,FactCheck2018}.
The next step is to use machine learning on these textual evidence snippets to evaluate a given assertion. In HybridFC, we refrain from using the machine learning module of text-based approaches. Instead, we compute an ordering for the list of text snippets returned by text-based approaches. To this end, we first determine the PageRank scores for all articles in the reference corpus~\cite{page1999pagerank} and select evidence sentences.  
Our evidence sentence selection module is based on the following hypothesis: "Documents (websites) with higher PageRank score provide better evidence sentences". Ergo, once provided with scored text snippets by a text-based approach, we select the top-$k$ evidence sentences coming from documents with top-$k$ PageRank scores. 
To each text snippet, we assign the PageRank score of its source article. Then, we sort the list of text snippets and use the $k$ snippets with the highest PageRank score.

We convert each of the selected snippets $t_i$ into a continuous vector representation using a sentence embedding model.  
We concatenate these sentence embeddings along with the trustworthiness scores~\cite{Nakamura2007} of their respective sources to create a single vector $\varphi_{\aleph}$. 
In short:

\begin{equation}
\varphi_{\aleph} = \bigoplus_{i=1}^k \left(b(t_i) \oplus \tau_i\right),
\end{equation}
where $\oplus$ stands for the concatenation of vectors, $b(t_i)$ is the sentence embeddings of $t_i$ and $\tau_i$ is the trustworthiness score of $t_i$. 
Our approach can make use of any text-based fact-checking approach that provides text snippets and a trustworthiness score, and allows us to compute PageRank score. Moreover, we can use any sentence embedding model.
For our experiments, we adapt the state-of-the-art text-based approach FactCheck~\cite{FactCheck2018} as a text-based fact checking approach, and make use of a pre-trained SBert Transformer model for sentence embeddings~\cite{sbert2019Reimers}.

\subsection{Path-based Component}
Path-based approaches determine the veracity of a given assertion by finding evidence paths in a reference KG.  
Our path-based component can make use of any existing path-based approach that takes the given assertion as input together with the reference KG and creates a single veracity score $\zeta$ as output.
This veracity score is the result of our path-based component.
Within our experiments, we use the state-of-the-art unsupervised path-based approach COPAAL~\cite{coopal22019}.

\subsection{KG Embedding-based Component}
KG embedding-based approaches generate a continuous representation of a KG using a mapping function. Based on a given KG embedding model, we create an embedding vector for a given assertion $(s,p,o)$ by concatenating the embedding of its elements and define the embedding mapping function for assertions $\varphi((s,p,o))$ as follows:    

\begin{equation}
\varphi((s,p,o)) = {\varphi}(s) \oplus {\varphi}(p) \oplus {\varphi}(o).
\end{equation}

In our approach, we can make use of any  KG embedding approach that returns both entities and relations embeddings. However, only a few approaches provide pre-trained embeddings for  large-scale KGs~(e.g., DBpedia). We use all approaches that provide pre-trained embeddings for DBpedia entities and relations in our experiments.

\begin{figure}[tb]
\centering
\includegraphics[width=.8\textwidth]{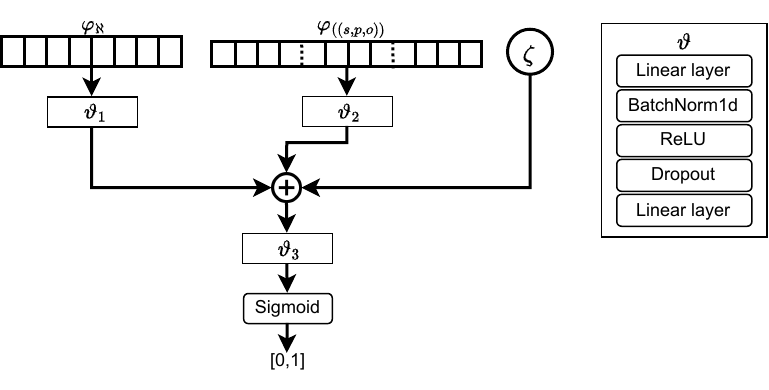}
\caption{\uq{Left: Overview of the architecture of HybridFC's neural network component. Right: Every $\vartheta_{i}$ is a multi-layer perceptron module.} }
\label{fig:arch1}
\end{figure}

\subsection{Neural Network Component}
The output of the three components above is the input to our neural network component. As depicted in Figure~\ref{fig:arch1}, the neural network component consists of three multi-layer perceptron modules that we name $\vartheta_{i}$.\footnote{During a first evaluation a simpler approach with only one multi-layer perceptron module (i.e., without $\vartheta_{1}$ and $\vartheta_{2}$) showed an insufficient performance.} 
Each of these modules consists of a \texttt{Linear} layer, a \texttt{Batch Normalization} layer, a \texttt{ReLU} layer, a \texttt{Dropout} layer and a final \texttt{Linear} layer. The output of the text-based component $\varphi_{\aleph}$ is fed as input to the first module. The output of the KG embedding-based component $\varphi((s,p,o))$ is fed to the second module. The output of the 2 modules and the veracity score $\zeta$ of the path-based component are concatenated and fed to the third module. The result of the third module is used as input to a sigmoid function $\sigma$, which produces a final output in the range $[0,1]$.
The calculation of the final veracity $\omega$ score for the given assertion can be formalized as follows:
\begin{equation}
\omega = \sigma \left( w_{\sigma}^{T} \vartheta_{3} \left( \vartheta_{1}(\varphi_{\aleph}) \oplus \vartheta_{2}(\varphi((s,p,o))) \oplus \zeta \right) \right)\,,
\end{equation}
where $w_{\sigma}$ is a weight vector that is multiplied with the output vector of the third module. Each of the three multi-layer perceptron modules~($\vartheta_{i}$) is defined as follows for an input vector $x$:
\begin{equation}
    \vartheta_{i} = W_{5,i}\times D_p(ReLU(W_{3,i}\times (BN(W_{1,i}\times x))))\,, \\
\end{equation}
where $x$ is an input vector, $W_{j,i}$ is the weight matrix of an affine transformation in the $j$-th layer of the multi-layer perceptron, $\times$ represents the matrix multiplication, \texttt{ReLU} is an activation function, $D_p$ stands for a \texttt{Dropout} layer~\cite{dropout2018watt}, and $BN$ represents the \texttt{Batch Normalization}~\cite{batchnrom2015sergey}. The latter is defined in the following equation:
\begin{equation}
    BN(x^\prime)=\beta + \gamma\frac{x^\prime-\mathrm{E}\left[x^\prime\right]}{\sqrt{\operatorname{Var}\left[x^\prime\right]}}\,,
\end{equation}
where, $x^\prime$ is the output vector of the first \texttt{Linear} layer and the input to the \texttt{Batch Normalization}, and $\mathrm{E}\left[x^\prime\right]$ and $\operatorname{Var}\left[x^\prime\right]$ are the expected value and variance of $x^\prime$, respectively. $\beta$ and $\gamma$ are weight vectors, which are learned during the training process via backpropagation to increase the accuracy~\cite{batchnrom2015sergey}. 
Furthermore, given the output of the \texttt{Linear} layer $x$ as input to the \texttt{Dropout} layer $D_p$, the output $\bar{x}$ is computed as:
\begin{equation}
\begin{cases}\bar{x} &= D_p(x)\\
\bar{x}_i& = \delta_ix_i\,,
\end{cases}
\end{equation}
where each $\delta_i$ follows the Bernoulli distribution of parameter $p$, i.e., $\delta$ is 1 with probability $p$, and 0 otherwise.


\section{Experimental Setup}
\label{sec:evaluation}
We evaluate our approach by comparing it with seven state-of-the-art fact-checking approaches. In the following, we first describe the datasets we rely upon. Then, we describe our experimental setting. 
\subsection{Datasets}
\subsubsection{Fact-checking Datasets.}

In our experiments, we use two recent fact-checking datasets that are often used in the literature~\cite{FactCheck2018,defacto2015,COPAAL2019}:  FactBench and BirthPlace/DeathPlace (BD). We use these datasets because they comprise entities of DBpedia, which is \begin{inparaenum}[(i)] 
\item large, and 
\item for which multiple pre-trained embedding models are available.
\end{inparaenum} 

We only use a subset of the original FactBench dataset because it was created in 2014, and is based on DBpedia version $3.9$~\cite{defacto2015}. Ergo, some of the facts it contains are outdated. For example, $(\texttt{:B.Obama}, \texttt{:presidentOf}, \texttt{:USA})$ was a correct assertion when the benchmark was created but is currently incorrect~(without the date information). 
We performed the following list of changes to obtain the benchmark used herein:
\begin{compactitem}
    \item We removed the date category from wrong assertions.
    \item We removed all assertions with Freebase entities. 
    \item We removed the $:team$ predicate, because there were many false positives in this category of assertions, since nearly all players changed their teams meanwhile.
\end{compactitem}
Our second evaluation dataset, dubbed BirthPlace/DeathPlace (short DB)~\cite{FactCheck2018}, aims to overcome a limitation of the FactBench dataset. 
It only contains assertions pertaining to birth and death places. The dataset was created based on the observation that some fact-checking approaches only check if the subject and object have a relation to each other while the type of the relation, i.e., whether it matches the property of the given assertion, is not always taken into account. Hence, all subjects and objects within the BD dataset have a relation to each other. This ensures that an approach only performs well on this dataset if it takes the type of the relation in assertions into account. 

\begin{table}[tb]
\centering
\caption{Overview of all correct facts used in our experiments. The train and test sets (train/test) are from the 2 benchmark datasets FactBench and BD from~\cite{FactCheck2018}.}
\begin{tabular}{llccl}
\toprule 

&\text {Property} & \text { |Sub|  } & \text { |Obj| }  & \text { Comment } \\
\midrule 
\parbox[t]{2.8mm} {\multirow{8}{*}{\rotatebox[origin=c]{90}{\textbf{FactBench}}}}

&\texttt { :birthPlace } & 75/75 &  67/65    & birth place (city) \\
 &\texttt { :deathPlace } & 75/75 & 54/48    & death place  (city)  \\
& \texttt { :award } & 75/75 & 5/5      & Winners of nobel prizes  \\
& \texttt { :foundationPlace } & 75/75 & 59/62   & Foundation place and timeof software companies  \\
& \texttt { :author } & 75/75 & 75/73   & Authors of science fiction books (one book/author) \\
& \texttt { :spouse } & 74/74 & 74/74   & Marriages between actors(after 2013/01/01) \\
& \texttt { :starring } & 22/21 & 74/74 & Actors starring in a movie \\
& \texttt { :subsidiary } & 54/50 & 75/75   & Company acquisitions \\
 \midrule
 \parbox[t]{2.8mm} {\multirow{2}{*}{\rotatebox[origin=c]{90}{\textbf{BD}}}}
 &\texttt { :birthPlace } & 51/52 &  45/35    & birth place (city) \\
 &\texttt { :deathPlace } & 52/51 & 42/38    & death place  (city)  \\

\bottomrule
\end{tabular}
\label{factbech-correct-dataset}
\end{table}

\begin{table}[tb]

\centering
\caption{Overview of the number of wrong assertions in the different categories of the train and test set (train/test) from the 2 benchmark datasets FactBench and BD~\cite{FactCheck2018}.}
\begin{tabular}{llcl}
\toprule 
&\text {Category} & \text { |Assertions| } & \text { Comment } \\
\midrule 
\parbox[t]{2.8mm} {\multirow{6}{*}{\rotatebox[origin=c]{90}{\textbf{FactBench}}}}

&\text { Domain } & 1000/985   & Replacing $s$ with another entity in the domain of $p$ \\
 &\text { Range } & 999/985   & Replacing $o$ with another entity in the range of $p$  \\
 &\text { DomainRange } & 990/989  &    Replacing $s$ or $o$ based on the domain and range of $p$, resp. \\
 &\text { Property } & 1032/997   & Replacing $s$ and $o$ based on $p$ connectivity  \\
 &\text { Random } & 1061/1031   & Randomly replacing $o$ or $s$ with other entities \\
 &\text { Mix } & 1025/1024  & Mixture of above categories \\

\midrule
 \parbox[t]{2.8mm} {\multirow{1}{*}{\rotatebox[origin=c]{90}{\textbf{BD}}}}
 &\texttt { type-based } & 206/206    & Replacing $s$ or $o$ of different RDF type \\

\bottomrule
\end{tabular} 
\label{factbech-wrong-dataset}
\end{table}

An overview of the two benchmarking datasets used in our evaluation in terms of the number of true and false assertions in training and testing sets, predicates, and some details about the generation of those assertions are presented in Tables~\ref{factbech-correct-dataset} and \ref{factbech-wrong-dataset}.
Note that both datasets were designed to be class-balanced. Hence, we do not need to apply any method to alleviate potential class imbalances in the training and test data.
However, we want to point out that the BD dataset provides less training examples than FactBench.

\subsubsection{Reference Corpus.}
Our text-based component makes use of a reference corpus. We created this corpus by extracting the plain text snippets from all English Wikipedia articles and loading them into an Elasticsearch instance. We used the dump from March 7th, 2022. For the Elasticsearch\footnote{\url{https://www.elastic.co/}} index, we used a cluster of 3 nodes with a combined storage of 1~TB and 32~GB RAM per node. 

\subsection{Evaluation Metric}

As suggested in the literature, we use the area under the receiver operator characteristic curve (AUROC) to compare the fact-checking results~\cite{unsupervised2020rule,COPAAL2019,FactCheck2018}. We compute this score using the knowledge-base curation branch of the GERBIL framework~\cite{paulheim2018swc,ngonga2019swc}.

\subsection{Setup Details and Reproducibility}

Within the sentence embedding module, we use a pre-trained SBert model.\footnote{We ran experiments with all available pre-trained models~(not shown in the paper due to space limitations) from the SBert homepage (\url{https://www.sbert.net/docs/pretrained_models.html}) and found that \texttt{nq-distilbert-base-v1} worked best for our approach.} Furthermore, we set $k=3$ in the sentence selection module. The size of the sentence embedding vectors generated by SBert is $768$, and the trustworthiness score values against each sentence vector, which leads to $|\varphi_{\aleph}|=(3 \times 768)+3=2307$. 

We use embeddings from 5 KG embedding models, where pre-trained DBpedia embeddings are available~\footnote{\uq{A large number of KG embedding algorithms~\cite{conexdemir2021convolutional,complextrouillon2016complex,daSilva2021} has been developed in recent years. However, while many of them show promising effectiveness, their scalability is often limited. For many of them, generating embedding models for the whole DBpedia is impractical (runtimes > 1 month). Hence, we only considered the approaches for which pre-trained DBpedia embeddings are available.}}. These models include: TransE~\cite{transE2013}, ConEx~\cite{conexdemir2021convolutional}, QMult~\cite{qmultdemir2021hyperconvolutional}, ComplEx~\cite{complextrouillon2016complex}, and RDF2Vec~\cite{rdf2vec2016ristoski}. For the FactBench dataset, we do not include experiments using RDF2Vec embeddings, because these embeddings were generated using a different version of DBpedia~(i.e., 2015-10) and missing embeddings of multiple entities (i.e., $40/1800$).\footnote{Fair comparison could not be possible with missing entities, which constitute many assertions.} However, we included RDF2Vec embedding in the BD dataset comparison. 
Different KG embedding models provide embedding vectors with different lengths. For example, the TransE model used within our experiment maps each entity and each relation to a vector with $100$ dimensions. This leads to a total size for $\varphi_{(s,p,o)}$ of $300$.


We use the Binary Cross Entropy (BCE) as loss function for training our neural network component.
We set the maximum number of epochs to 1000 with a batch size of 1/3 of the training data size.
The training may have to be stopped earlier in case the neural network component starts to overfit.
To this end, we calculate the validation loss every 10th epoch and if this loss does not decrease for 50 epochs, the training is stopped.

All experiments are conducted on a machine with 32 CPU cores, 128~GB RAM and an NVIDIA GeForce RTX 3090.
We provide hyperparameter optimization, training, and evaluation scripts on our project page for the sake of reproducibility.

\subsection{Competing Approaches}
We compare HybridFC in different configurations to FactCheck~\cite{FactCheck2018}, COPAAL~\cite{COPAAL2019}, and KV-Rule~\cite{unsupervised2020rule}, which are the state-of-the-art approaches of the text-, path- and rule-based categories, respectively.
We also compare our results to those four KG embedding-based approaches for which pre-trained DBpedia embedding models are available. 
We employ these models for fact checking by training the neural network module $\vartheta_{2}$ of our approach based only on the output of the KG-based component. The output of this neural network module is then directly used as input for the final sigmoid function.
We do not compare our results with results of the hybrid approaches mentioned in Section~\ref{sec:motivationRelatedWork} because ExFaKT and Tracy mainly focus on generating human-comprehensible explanations and do not produce the veracity score, and FACTY focuses on calculating the veracity of assertions containing long-tail vertices~(i.e., entities from less popular domains, for example, cheese varieties). 

\section{Results and Discussion}
\label{sec:results}
Tables~\ref{auc-train-dataset} and~\ref{auc-test-dataset} show the AUROC scores for the different hybrid and competing approaches on the FactBench train and test datasets, respectively. We can see that HybridFC performs best when it uses the TransE embedding model. \uq{This is not unexpected as TransE is one of the simplest embedding models that supports property composition: Given two properties $p_1$ and $p_2$, TransE entails that $\varphi(p_1 \circ p_2) \approx \varphi(p_1) + \varphi(p_2)$.  With TransE as its embeddings model, HybridFC significantly outperforms all competing approaches on the test data.}\footnote{We use a Wilcoxon signed rank test with a significance threshold $\alpha=0.05$.}
\begin{table}[h!tb]
\centering
\caption{Area under the curve (AUROC) score on different categories of FactBench train sets. T stands for text-based approach, P for path-based approach, R for rule-based approaches, and KG-emb for KG-embedding-based approaches.}

\begin{tabular}{clccccccc}
\toprule 
&& \text { Domain } & \text { Range } & \text { DomainRange } & \text { Mix } & \text { Random } & \text { Property } & \text { Avrg. } \\
\midrule 
\parbox[t]{2.8mm} {\multirow{1}{*}{\rotatebox[origin=c]{90}{\textbf{T}}}}
& \text { FactCheck~\cite{FactCheck2018}} & 0.69 & 0.69 & 0.68 & 0.65 & 0.68 & 0.57 & 0.66 \\
\midrule
 \parbox[t]{2.8mm} {\multirow{1}{*}{\rotatebox[origin=c]{90}{\textbf{P}}}}
 &\text { COPAAL~\cite{COPAAL2019} } & 0.67 & 0.67 & 0.68 & 0.65 & 0.69 & 0.68 & 0.67  \\
 \midrule
 \parbox[t]{2.8mm} {\multirow{1}{*}{\rotatebox[origin=c]{90}{\textbf{R}}}}
 &\text { KV-Rule~\cite{unsupervised2020rule} } & 0.57 & 0.57 & 0.58 & 0.58 & 0.63 & 0.63 & 0.59 \\
 \midrule
  \parbox[t]{2.8mm} {\multirow{4}{*}{\rotatebox[origin=c]{90}{\textbf{KG-emb}}}}
& \text { TransE~\cite{transE2013}} & 0.67 & 0.61 & 0.78 & 0.66 & 0.92 & 0.97 & 0.76 \\
&  \text { ConEx~\cite{conexdemir2021convolutional} } & 0.64 & 0.67 & 0.68 & 0.86 & 0.96 & 0.88 &  0.78 \\
&  \text { ComplEx~\cite{complextrouillon2016complex}} & 0.78 & 0.66 & 0.74 & 0.80 & 0.98 & 0.97 &  0.82\\
& \text { QMult~\cite{qmultdemir2021hyperconvolutional} } & 0.83 & 0.73 & 0.75 & 0.86 & 0.97 & 0.98 & 0.85 \\
\midrule
\parbox[t]{2.8mm} {\multirow{4}{*}{\rotatebox[origin=c]{90}{\textbf{HybridFC}}}}
&\text{TransE} & \textbf{0.94} & \textbf{0.94} & \textbf{0.96} & \textbf{0.90} & \textbf{0.99} & \textbf{0.99} & \textbf{0.95}\\
 &\text{ConEx} & 0.81 & 0.79 & 0.81 & 0.74 & 0.82 & 0.80 &  0.79\\
 &\text{ComplEx} & 0.94 & 0.94 & 0.94 & 0.86 & 0.95 & 0.97 & 0.93\\
 &\text{QMult} & 0.90 & 0.89 & 0.89 & 0.81 & 0.91 & 0.94 & 0.89\\
\bottomrule
\end{tabular}
\label{auc-train-dataset}
\end{table}

\begin{table}[h!tb]
\centering
\caption{Area under the curve (AUROC) score on different categories of FactBench test sets; the abbreviations are: T/Text-based approaches, P/Path-based approaches, R/Rule-based approaches, and KG-emb/KG embedding-based approaches.}
\begin{tabular}{clccccccc}
\toprule 
&& \text { Domain } & \text { Range } & \text { DomainRange } & \text { Mix } & \text { Random } & \text { Property } & \text { Avrg. } \\
\midrule 
\parbox[t]{2.8mm} {\multirow{1}{*}{\rotatebox[origin=c]{90}{\textbf{T}}}}

& \text { FactCheck~\cite{FactCheck2018}} & 0.67 & 0.67 & 0.66 & 0.61 & 0.66 & 0.59 & 0.64 \\
 \midrule
 
 \parbox[t]{2.8mm} {\multirow{1}{*}{\rotatebox[origin=c]{90}{\textbf{P}}}}

& \text { COPAAL~\cite{COPAAL2019} } & 0.67 & 0.68 & 0.68 & 0.65 & 0.69 & 0.69 & 0.68 \\
 \midrule
 \parbox[t]{2.8mm} {\multirow{1}{*}{\rotatebox[origin=c]{90}{\textbf{R}}}}
 & \text { KV-Rule~\cite{unsupervised2020rule} } & 0.57 & 0.57 & 0.57 & 0.58 & 0.61 & 0.62 & 0.59 \\
 \midrule
  \parbox[t]{2.8mm} {\multirow{4}{*}{\rotatebox[origin=c]{90}{\textbf{KG-emb}}}}

& \text { TransE~\cite{transE2013}} & 0.63 & 0.60 & 0.63 & 0.64 & 0.87 & 0.96 & 0.72 \\
&  \text { ConEx~\cite{conexdemir2021convolutional}} & 0.50 & 0.50 & 0.50 & 0.52 & 0.60 & 0.60 &  0.54 \\
&  \text { ComplEx~\cite{complextrouillon2016complex}} & 0.58 & 0.58 & 0.52 & 0.62 & 0.86 & 0.95 &  0.69\\
& \text { QMult~\cite{qmultdemir2021hyperconvolutional}} & 0.57 & 0.62 & 0.55 & 0.69 & 0.84 & 0.93 & 0.70 \\
\midrule
\parbox[t]{2.8mm} {\multirow{4}{*}{\rotatebox[origin=c]{90}{\textbf{HybridFC}}}}

& \text { TransE } & \textbf{0.80} & \textbf{0.80} & \textbf{0.81} & \textbf{0.78} & \textbf{0.95} & \textbf{0.99} & \textbf{0.86}\\
& \text { ConEx } & 0.77 & 0.78 & 0.79 & 0.71 & 0.80 & 0.70 &  0.75\\
& \text { ComplEx } & 0.75 & 0.76 & 0.74 & 0.72 & 0.93 & 0.97 & 0.81\\
& \text { QMult } & 0.69 & 0.73 & 0.71 & 0.69 & 0.91 & 0.94 & 0.77\\
\bottomrule
\end{tabular}
\label{auc-test-dataset}
\end{table}

\begin{table*}[!htb]
\centering
\caption{Area under the curve (AUROC) scores on the BD dataset; the abbreviations are: T stands for text-based approaches, P for path-based approaches, R for rule-based approaches, KG-emb for KG-embedding-based approaches.}
\footnotesize
\setlength{\tabcolsep}{4pt}
\begin{tabular}{@{}lccccccccccccc@{}}
  \toprule
  &\textbf{T}
  &\textbf{P}
  &\textbf{R}
  &\multicolumn{5}{c}{\textbf{KG-emb}}&\multicolumn{5}{c}{\textbf{HybridFC}}\\
  	\cmidrule(lr){2-2} \cmidrule(lr){3-3} \cmidrule(lr){4-4}
  	\cmidrule(lr){5-9} \cmidrule(lr){10-14} 
    &
 \rotatebox[origin=l]{90}{FactCheck~\cite{FactCheck2018}} &      
 {\rotatebox[origin=l]{90}{COPAAL~\cite{COPAAL2019}}} &     
 {\rotatebox[origin=l]{90}{KV-Rule~\cite{unsupervised2020rule}}} &
 \rotatebox[origin=l]{90}{TransE~\cite{transE2013}} &
 {\rotatebox[origin=l]{90}{ConEx~\cite{conexdemir2021convolutional}}} &     
 {\rotatebox[origin=l]{90}{ComplEx~\cite{complextrouillon2016complex}}} &     
 {\rotatebox[origin=l]{90}{QMult~\cite{qmultdemir2021hyperconvolutional}}}  &     
 {\rotatebox[origin=l]{90}{RDF2Vec~\cite{rdf2vec2016ristoski}}} &     
 {\rotatebox[origin=l]{90}{TransE}} &     
 {\rotatebox[origin=l]{90}{ConEx}}  &     
 {\rotatebox[origin=l]{90}{ComplEx}} &     
 {\rotatebox[origin=l]{90}{QMult}} &     
 {\rotatebox[origin=l]{90}{RDF2Vec}}  \\ \midrule
	  \textbf{Train} &  
	  {0.51} &  {0.67} &  {0.76} &{0.69} & {0.50}  &  {0.73}  & {0.60} & {0.67}&{0.80} & {0.51}  &  {0.74}  & {0.60} & {0.74}  \\
	 \textbf{Test}    &  
	{0.49} &  {0.70} &  \textbf{0.81}  & {0.54} &   {0.50} &  {0.54} & {0.55} & {0.62}&{0.69} & {0.50}  &  {0.57}  & {0.58} & {0.68} \\
	  \bottomrule
\end{tabular}
\label{auc-bpdp-dataset}
\end{table*}

Note that FactCheck does not achieve the performance reported in~\cite{FactCheck2018} within our evaluation. This is due to \begin{inparaenum}[(i)] \item the use of a different English Wikipedia as reference corpus---Syed et al. showed that they achieve better results with the larger ClueWeb corpus---and \item the fact that we had to remove triples from the FactBench dataset.
\end{inparaenum}

The overall performance of COPAAL is better than the performance of FactCheck, ConEx, QMult and KV-Rule on the test set. However, we observe large performance differences with respect to the different properties.
While COPAAL achieves the second best AUROC scores after HybridFC for 
6 out of the 8 properties it struggles to achieve good results for \texttt{:award} and \texttt{:author}. 
These experimental results suggest that our approach makes good use of the diversity of the performance of the approaches it includes. In particular, it seems to rely on COPAAL's good performance on most of the properties while being able to complement  COPAAL's predictions with that of other algorithms for properties on which COPAAL does not perform well. 

On the BD dataset, KV-rule outperforms all other approaches on the test split. COPAAL achieves the second best score, closely followed by the TransE-based HybridFC variant.
The results confirm that the unsupervised fact-checking approaches COPAAL and KV-rule achieve good results for the \texttt{:birthPlace} and \texttt{:deathplace} properties.
A closer look at the results reveals two main reasons for the lower result of the TransE-based HybridFC variant on the test dataset. First, FactCheck fails to extract pieces of evidence for most of the assertions. Second, FactCheck, the embedding-based approaches as well as the HybridFC variants are supervised approaches and suffer from the small size of the train split of the BD dataset. This is confirmed by our observation that the neural network component tends to overfit during the training phase.
\section{Ablation Study}
\label{sec:ablation}

\begin{table}[tb]
\centering
\caption{Results of our ablation study on the FactBench test set and BD dataset. D stands for Domain, R for Range, DR for DomainRange, Ran. for Random, Prop. for Property, and Avg. for average. TC stands for text-based component, PC for path-based component, EC for embedding-based component, and  the symbol $+$ indicates the combination of $2$ components. Best performances are bold, second-best are underlined.}
\begin{subtable}[t]{0.60\textwidth}
\centering
\caption{FactBench test set}
\label{auc-factbench-dataset-ablation}
\begin{tabular}{clccccccc}
\toprule 
& \text {D} & \text {R} & \text {DR} & \text { Mix } & \text { Ran. } & \text { Prop. } & \text { Avg. } \\
\midrule 
 \uq{\text { TC }} &  0.76 & 0.77 & 0.76 & 0.69 & 0.77 & 0.64 & 0.73  \\
  \uq{\text { PC }} & 0.68 & 0.69 & 0.69 & 0.65 & 0.70 & 0.69 & 0.68  \\
  \uq{\text { EC }} &  0.63 & 0.61 & 0.62 & 0.64 & 0.86 & 0.97 & 0.72  \\
 \uq{\text { TC+EC}} &  0.76 & \underline{0.78} & 0.76 & \underline{0.74} & \underline{0.92} & \underline{0.98} & \underline{0.82}  \\
\uq{\text { TC+PC}} &  \underline{0.77} & 0.77 & \underline{0.77} & 0.7 & 0.79 & 0.67 & 0.74  \\
\uq{\text { PC+EC }} & 0.71 & 0.7 & 0.69 & 0.72 & 0.89 & 0.97 & 0.78 \\ \midrule 
 \text {HybridFC} & \textbf{0.80} & \textbf{0.80} & \textbf{0.81} & \textbf{0.78} & \textbf{0.95} & \textbf{0.99} & \textbf{0.86}\\
\bottomrule
\end{tabular}
\end{subtable}
     \hfill
    \begin{subtable}[t]{0.35\textwidth}
\centering
\caption{BD dataset}
\label{auc-bpdp-dataset-ablation}
\begin{tabular}{clll}
\toprule 
& \text { Train } & \text { Test } \\
\midrule
 
 \uq{\text { TC }} & 0.59 & 0.56 \\
 \uq{\text { PC }} & 0.67 & \textbf{0.70} \\
 \uq{\text { EC }} & 0.69 & 0.56 \\
 \uq{\text { TC+EC}} & \underline{0.79} & 0.65 \\
 \uq{\text { TC+PC }} & 0.67 & 0.64 \\
 \uq{\text { PC+EC }} & 0.74 & 0.66 \\ \midrule
 \text{HybridFC} & \textbf{0.80} & \underline{0.69} \\
\bottomrule
\end{tabular}
    \end{subtable}
\label{auc-test-dataset-ablation}
\end{table}

\uq{Our previous experiments suggest that HybridFC performs best in combination with TransE. Hence, we use it as default setting throughout the rest of the paper and overload HybridFC to mean HybridFC with TransE embeddings. To evaluate the contribution of the different components of HybridFC to its performance, we rerun our evaluation for each component~(i.e., text-based (TC), path-based (PC), and embedding-based (EC)) individually and as pairwise combination of different components (TC+PC, TC+EC, PC+EC). 
The results for the FactBench test and the BD datasets are shown in Tables~\ref{auc-factbench-dataset-ablation} and~\ref{auc-bpdp-dataset-ablation}.\footnote{Due to space limitation we exclude the results of FactBench train set. These results are available on our GitHub page.} 
The results suggest that the individual path-based and embedding-based components achieve results similar to those of COPAAL and TransE, respectively. Our text-based component achieves better results than FactCheck.
On the FactBench test datasets, the combination of two components leads to better results than the single components. Similarly, HybridFC, i.e., the combination of all three components, leads to significantly better results than all pairwise combinations, where significance is measured using a Wilcoxon signed rank test with a p-value threshold of 0.05. Here, our null hypothesis is that the performances of the approaches compared are sampled from the same distribution. 
For the BD dataset, the pairwise combinations of components suffer from the same overfitting problem as HybridFC.
Overall, our results in Table~\ref{auc-factbench-dataset-ablation} suggest that our text component commonly achieves the highest average performance on datasets that provide enough training data.
The text component is best supplemented by the embedding-based component. HybridFC outperforming all combinations of two components on FactBench suggests that in cases in which HybridFC is trained with enough training data, each of the three components contributes to the better overall performance of HybridFC.}

\section{Conclusion}
\label{sec:conclusion}
In this paper, we propose HybridFC--a hybrid fact-checking approach for KGs. HybridFC aims to alleviate the problem of manual feature engineering in text-based approaches, cases in which paths between subjects and objects are unavailable to path-based approaches, and the poor performance of pure KG-embedding-based approaches by combining these three categories of approaches.
We compare HybridFC to the state of the art in fact checking for KGs. Our experiments show that our hybrid approach is able to outperform competing approaches in the majority of cases.
As future work, we will exploit the modularity of HybridFC by integrating rule-based approaches. 
We also plan to explore other possibilities to select the best evidence sentences.



\section*{Supplemental Material Statement} 
\begin{compactitem}
\item The source code of HybridFC, the scripts to recreate the full experimental setup, and the required libraries can be found on GitHub.\footnote{Source code: \url{https://github.com/dice-group/HybridFC}} 
\item Datasets used in this paper and the output generated by text-based and path-based approaches on these datasets are available at Zenodo~\cite{dataset1}.  
\item Pre-trained embeddings for these datasets are also available at Zenodo~\cite{dataset2}.
\end{compactitem}

\section*{Acknowledgments}
The work has been supported by the EU H2020 Marie Skłodowska-Curie project KnowGraphs (no. \ 860801), the German Federal Ministry for Economic Affairs and Climate Action (BMWK) funded project RAKI (no. \ 01MD19012B), and  the German Federal Ministry of Education and Research (BMBF) funded EuroStars projects 3DFed (no. \ 01QE2114B) and FROCKG (no. \ 01QE19418). We are also grateful to Daniel Vollmers and Caglar Demir for the valuable discussion on earlier drafts. This is the pre-print version of paper, which is accepted at ISWC 2022.

\bibliographystyle{splncs04}
\bibliography{finalpaper}




\end{document}